\documentclass[runningheads]{llncs}

\usepackage{eccv}

\usepackage{eccvabbrv}

\usepackage{graphicx}
\usepackage{booktabs}
\usepackage[accsupp]{axessibility}  

\usepackage[pagebackref,breaklinks,colorlinks,citecolor=eccvblue]{hyperref}

\usepackage{orcidlink}
\usepackage{multirow} 
\usepackage{comment}
\begin{document}

\title{AdaDistill: Adaptive Knowledge Distillation for Deep Face Recognition
} 

\titlerunning{AdaDistill}

\author{Fadi Boutros \inst{1} \and Vitomir Štruc \inst{2} \and Naser Damer \inst{1,3}
}

\authorrunning{Boutros et al.}

\institute{Fraunhofer Institute for Computer Graphics Research IGD, Darmstadt, Germany \and
University of Ljubljana, Ljubljana, Slovenia \and
Department of Computer Science, TU Darmstadt, Darmstadt, Germany\\
\email{fadi.boutros@igd.fraunhofer.de}}

\maketitle

\begin{abstract}
Knowledge distillation (KD) aims at improving the performance of a compact student model by distilling the knowledge from a high-performing teacher model. 
In this paper, we present an adaptive KD approach, namely AdaDistill, for deep face recognition. 
The proposed AdaDistill embeds the KD concept into the softmax loss by training the student using a margin penalty softmax loss with distilled class centers from the teacher. 
Being aware of the relatively low capacity of the compact student model, we propose to distill less complex knowledge at an early stage of training and more complex one at a later stage of training. 
This relative adjustment of the distilled knowledge is controlled by the progression of the learning capability of the student over the training iterations without the need to tune any hyper-parameters.
Extensive experiments and ablation studies show that AdaDistill can enhance the discriminative learning capability of the student and demonstrate superiority over various state-of-the-art competitors on several challenging benchmarks, such as IJB-B, IJB-C, and ICCV2021-MFR \footnote{\url{https://github.com/fdbtrs/AdaDistill}}.
\end{abstract}

\vspace{-9mm}
\section{Introduction}
\label{sec:intro}
\vspace{-3mm}
State-of-the-art (SOTA) face recognition (FR) models \cite{deng2019arcface,DBLP:conf/cvpr/WangWZJGZL018,DBLP:conf/cvpr/HuangWT0SLLH20} rely on training deep neural networks (DNN) \cite{ResNet} with millions of trainable parameters and high computational cost. However, deploying such models on mobile and edge devices remains a challenge due to the limited computational capacities of these devices \cite{DBLP:conf/iccvw/DengGZDLS19}. 
Designing efficient network architectures \cite{DBLP:conf/iccvw/YanZXZWS19,MFN} and compressing existing deep neural networks (DNNs) \cite{DBLP:conf/icpr/BoutrosDK22}, using e.g. parameter pruning or model quantization, are common approaches to mitigate this challenge. 
While these approaches have been shown to reduce the required computational resources \cite{DBLP:conf/iccvw/YanZXZWS19,MFN,DBLP:conf/icpr/BoutrosDK22}, they commonly lead to lower verification accuracies compared to the more computationally expensive SOTA models \cite{DBLP:conf/iccvw/DengGZDLS19}.

Knowledge distillation is an effective technique to improve the performance of compact models by transferring the knowledge from a large and well-performing teacher model to a comparably smaller student model \cite{KD,ReFO,TripletDistillation,EKD}.
With guidance from the teacher, the student can learn more effectively and achieve high accuracy, despite having fewer parameters.
The idea of KD was originally proposed by Bucilua et al. \cite{DBLP:conf/kdd/BucilaCN06} in 2006 and popularized in 2015 by Hinton et al. \cite{KD}. While it was initalliy applied to image classification \cite{KD}, it was  later widely adopted and advanced for different machine-learning tasks. In FR, for example, several works proposed to improve compact FR performances using KD \cite{ReFO,MarginKD,TripletDistillation,EKD,ShrinkTeaNet}.
From such works, \cite{TripletDistillation} opted to adapt and modify the traditional triplet loss in the distillation process to include a dynamic margin. Others opted to adopt a rank-based loss to optimize the distillation process based on pair verification performance \cite{EKD}. 
The feature direction and sample distributions transfer to the student through an angular loss was also recently proposed \cite{ShrinkTeaNet}. While \cite{ReFO} proposed a feature-based KD approach with reverse distillation and a student proxy, this required multiple phases of student-teacher training and imposed strict constraints on the student to imitate the teacher embeddings. 
On the other hand, \cite{MarginKD} fixed class centers learned by the teacher model and trained the student with a margin-based softmax loss. However, because these classes are fixed, this might not be optimal for all the training stages.
\begin{figure}[t!]
\vspace{-3mm}
     \centering
     \begin{subfigure}[b]{0.40\linewidth}
         \centering
         \includegraphics[width=\linewidth]{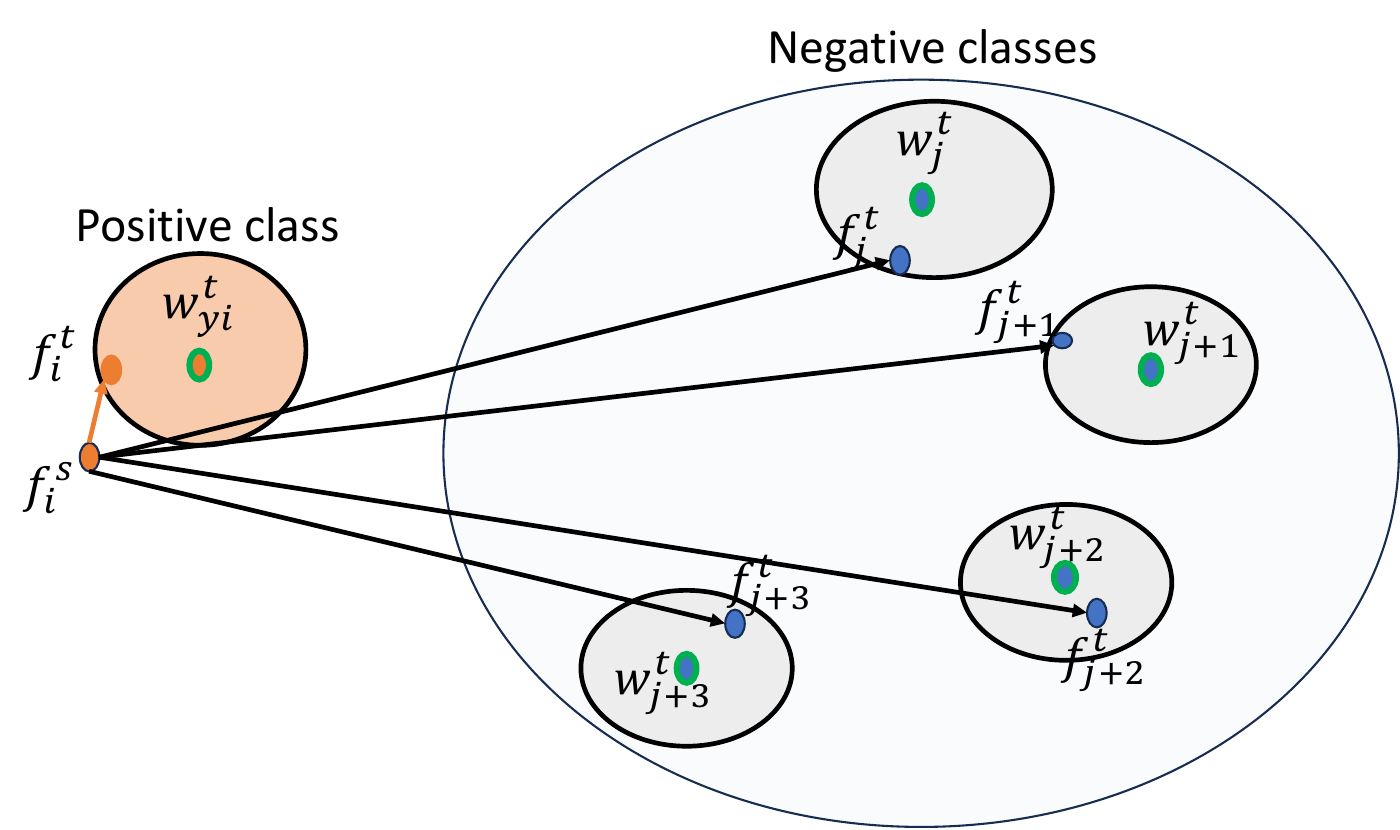}
         \caption{AdaDistill - Early stage
         }
         \label{fig:overview_intro_ealry}
     \end{subfigure}
     \begin{subfigure}[b]{0.40\linewidth}
         \centering
         \includegraphics[width=\linewidth]{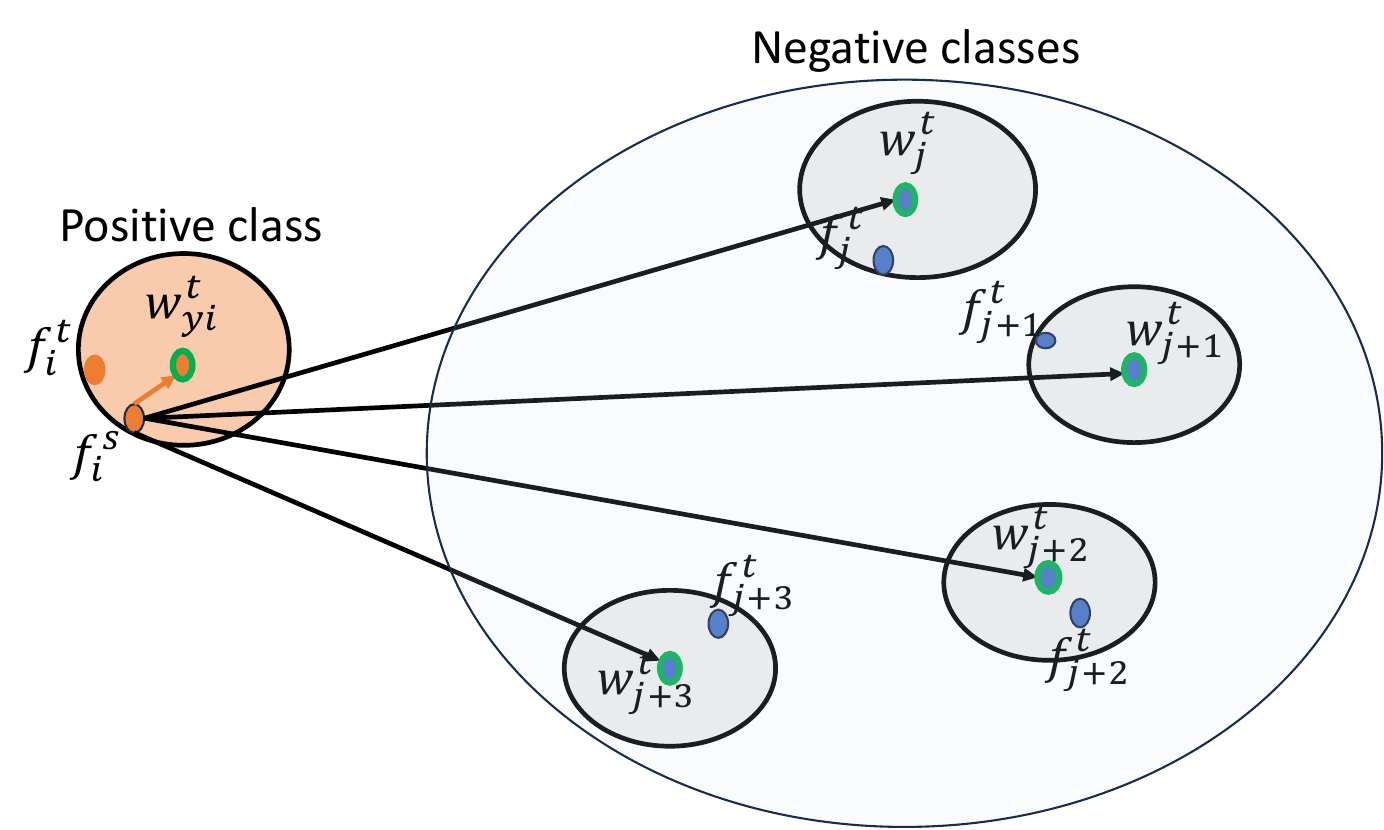}
         \caption{AdaDistill - Later stage
         }
         \label{fig:overview_intro_later}
     \end{subfigure}
     \vspace{-2mm}
        \caption{
        Overview of our adaptive KD approach. At an early stage of training, $f^s_i$ (from the student) is pushed to be close to the counterpart $f^t_i$ obtained from the teacher and far from other $f^t_j$ of different classes. As training goes on, $f^s_i$  is pushed to be close to its class center, $w_{yi}^t$ (distilled from the teacher) and far from all other class centers, $w_{j}^t$. 
       }
        \label{fig:overview_intro}
        \vspace{-8mm}
\end{figure}

In this study, we introduce a novel adaptive knowledge distillation approach termed AdaDistill. AdaDistill addresses several key challenges of previous approaches by eliminating the reliance on fixed class centers, avoiding the need for recursive training, and circumventing the need for multiple loss balance. Furthermore, it accounts for the student-teacher gap, ensuring effective knowledge transfer from the teacher to its student in an adaptive manner.
We embedded our adaptive distillation concept into the margin penalty softmax loss \cite{deng2019arcface}, aiming at maximizing the distance between features of different classes by incorporating a margin penalty between feature representations of each sample and its class center, which is distilled from the teacher. 
Additionally, we consider the capacity gap, in terms of learning capabilities, between the shallow student and large teacher models and propose to transfer less complex knowledge at an early stage of training and more complex one at a later stage of training. 


To achieve that, an adaptive procedure based on the exponential moving average over class-conditioned feature representations is introduced that allows for the distilling of knowledge at the early stages based on sample-to-sample comparisons, while at the later stages, more complex information captured by sample-to-class-center comparisons is exploited, as illustrated in Fig. \ref{fig:overview_intro_later}. 
%
%
This 
 comparison process evolves dynamically over the training stages and is
 determined by the student based on its ability to imitate feature representations from the teacher, i.e., using positive cosine similarity between feature representations from the teacher and the student of each sample.
We also incorporate the importance of hard samples in the class center estimation, as induced by the teacher.  
The proposed AdaDistill does not require manually tuning additional hyper-parameters other than the margin penalty value of the softmax loss, where class centers estimation and the importance of hard samples are dynamically adaptive over the training iteration. 
We present several ablation studies to support our AdaDistill design choices. We also conduct extensive evaluations on 10  benchmarks, demonstrating the superiority of our AdaDistill over SOTA KDs.

\vspace{-4mm}
\section{Related Works}
\label{sec:related_work}
\vspace{-3mm}
KD approaches in the literature can be categorized, based on the form of the distilled knowledge, under three categories, response-based, feature-based, and relation-based KD \cite{DBLP:journals/ijcv/GouYMT21}.
Response-based KD refers to mimicking the final output prediction between the student and the teacher.
The work by Hinton et al. \cite{KD} was one of the pioneer works proposing response-based KD that minimizes the Kullback-Leibler (KL) divergence between the soft class probabilities of the teacher and student models.
This inspired several research works, proposing KD approaches to transfer a diversity of knowledge forms from the teacher to the student beyond class probabilities.
Feature-based KDs, also known as instance-based \cite{ReFO},  utilizes instance-wise knowledge from intermediate feature representations as additional supervision to train the student model.
Within this category, FitNet \cite{FitNet} proposed to distill the knowledge from the intermediate feature maps by utilizing a regressor to match the dimensions of the intermediate feature maps between the teacher and the student. ShrinkTeaNet \cite{ShrinkTeaNet} presented an angular loss to transfer the feature direction and sample distributions from the teacher embedding space to its student. 
MarginDistillation \cite{MarginKD} trained a student model with margin-based softmax loss and fixed class centers learned by the teacher model.
TripletDistillation \cite{TripletDistillation} extended the traditional triplet loss used in deep face recognition by incorporating a dynamic margin. This dynamic margin is derived from the similarity structures among different identities from the teacher model. Very recently, Li et al. \cite{ReFO} proposed a feature-based KD approach with reverse distillation and a student proxy, aiming at minimizing the intrinsic gap between the teacher and the student model.
Unlike response-based and feature-based KD, which utilizes the output of a single layer or pairs of data samples, relation-based KD consideres the relation between several layers or data points within a batch of samples.
Here, DarkRank \cite{DarkRank} proposed to distill the relative similarity based on ranking between samples within each training batch.
CCKD \cite{CCKD} presented a KD approach to transfer the pairwise similarity on the instance level and the correlation between the instances on the batch level.
RKD \cite{RKD} proposed to transfer relations among data points learned by the teacher to the student model using distance-wise and angle-wise distillation losses.
EKD \cite{EKD} introduced a rank-based loss, aiming at selecting key pair-relations based on face verification metrics to be distilled to the student model.


Our proposed AdaDistill aims at distilling knowledge from a teacher to a student model by leveraging their respective feature representations from the embedding space. Thus, it is a feature-based method. 
Compared to ReFo \cite{ReFO}, which aims at training a student model to imitate exact teacher embedding space, our AdaDistill imposes less strict constraints on the student by pushing the samples to be relatively closer to their class center than other negative class centers.
Also, unlike ReFo, our AdaDistill does not require s student proxy or reverse distillation through multiple phases of student-teacher training. 
At an early stage of student training, our AdaDistill and ReFo \cite{ReFO} share similar learning objectives, while at the later stage of training, our AdaDistill and MarginDistillation \cite{MarginKD} share similar learning objectives.
Different from MarginDistillation \cite{MarginKD} that utilized fixed class centers from the teacher, our AdaDistill adapts the class center at different stages of the model training, enabling better convergence and higher verification accuracies. 
Compared to ShrinkTeaNet \cite{ShrinkTeaNet} that minimized the angular distance between the teacher and student embedding representations for each sample,  our AdaDistill transfers richer and more discriminative information from class centers learned by the teacher to its student.

Similarly to \cite{ReFO,MarginKD}, and differently from \cite{ShrinkTeaNet,EKD,RKD,FitNet,CCKD}, our AdaDistill utilizes only the distillation loss as the main training objective for the student without additional supervision from the classification loss.
This mitigates the challenge of simultaneously optimizing different learning objectives or balancing the values of interest (i.e., loss weighting terms).












\begin{figure}[t!]
\vspace{-2mm}
    \centering   \includegraphics[width=0.70\linewidth]{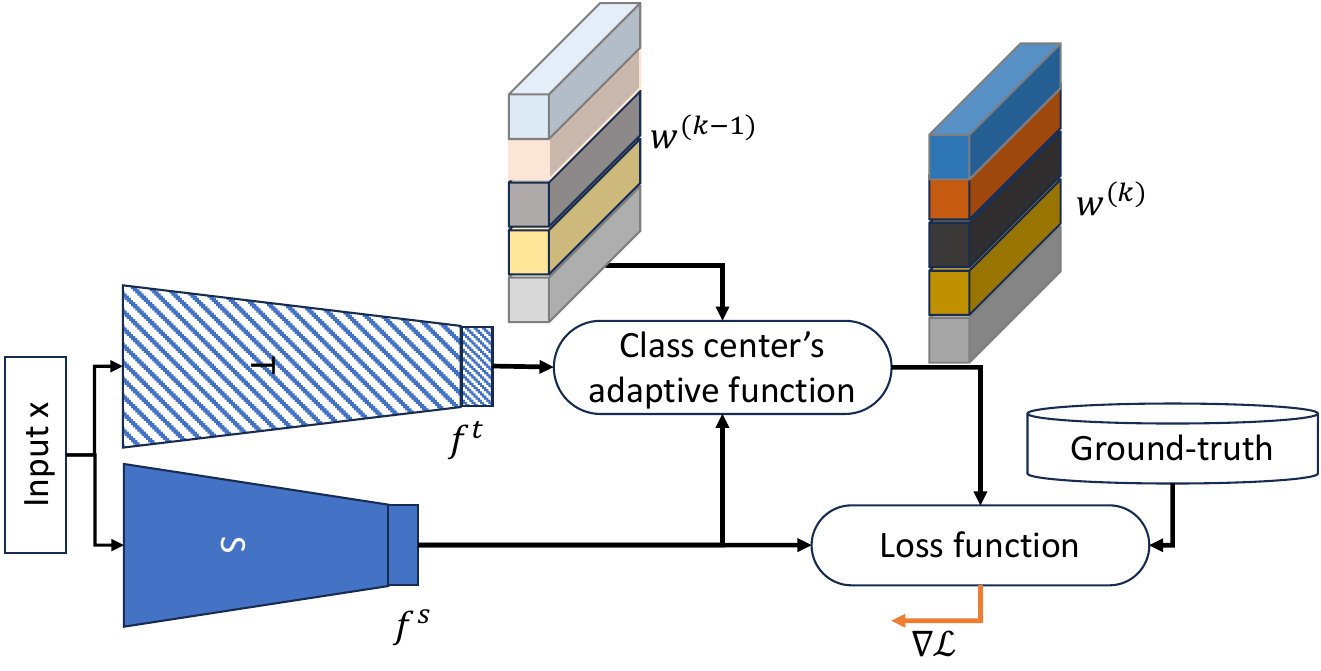}
    \vspace{-4mm}
    \caption{Overview of AdaDistill. A batch of samples $x$ from the  $k-th$ training iteration is passed to $T$ and $S$ to output two sets of feature representations $f^t$ and $f^s$, respectively.  Then, the class centers $w^{(k)}$ are calculated (Eq. \ref{eq:w}) based on the importance of hard samples and the learning capability of the student (Eq.\ref{eq:w_alpha}). Finally, the loss is calculated (Eq. \ref{eq:arcdistill}) and the weights of $S$ are updated.
    }
    \label{fig:overview}
    \vspace{-6mm}
\end{figure}
\vspace{-4mm}
\section{Methodology}
\vspace{-3mm}
This section first presents preliminaries on conventional feature-based KD, which we use as a baseline. Next, it presents the concept of template distillation through the Softmax loss. Finally, it introduces the proposed AdaDistill, which considers the gap between the student and the teacher, as well as the importance of hard samples over the distillation training. 
An overview of the proposed AdaDistill approach is presented in Fig. \ref{fig:overview}.

\vspace{-2mm}
\subsection{Preliminary: Knowledge distillation}
\vspace{-1mm}


Throughout the paper, we denote the teacher and student models as $T$ and $S$, respectively.  
The training dataset is denoted as $D$ and consists of training samples $x \in X$ and their corresponding class labels $y \in \{1,\ldots,c\}$, where $c$ is the number of classes. 
The feature representations of sample $x_i$, extracted by $T$ and $S$, are denoted as $f_i^t$ and $f_i^s$, respectively.

Given a pretrained teacher $T$, the general loss function for training $S$ in the KD paradigm can generally be formulated as:
\begin{equation}
\label{eq:general_kd}
    \mathcal{L}_{S}= \lambda \mathcal{L}_{main} + \beta \mathcal{L}_{KD},
\end{equation}
where $\mathcal{L}_{main}$ is the task-specific loss function, e.g.  Cross-entropy for classification learning, and $\mathcal{L}_{KD}$  is the KD loss (e.g., KL divergence) used to transfer the knowledge from $T$ to $S$.
$\lambda$ and $\beta$ are weightings parameters. Eq. \ref{eq:general_kd} represents a common loss for KD and $\lambda$ and $\beta$ do not necessarily sum up to one. Approaches, such as \cite{ReFO}, set $\lambda$ to zero, while others utilize both losses \cite{EKD}. 
Response-based KD approaches, such as \cite{KD}, utilize the KL divergence to match the output distributions, i.e., the classification output, of the teacher and student models.
However, in FR, matching the final output is less effective, as the goal of FR is learning to produce discriminative feature representations, instead of only accurate classification with the Softmax layer \cite{EKD,ReFO}. Therefore,  SOTA feature-based KD \cite{ReFO} aim to mimic intermediate feature representations, $f_i^t$ and $f_i^s$, instead of the final output probabilities. We follow these insights with AdaDistill, and, therefore, do not consider the KD term in the distillation process, i.e., $\beta=0$.  

Several previous works \cite{ReFO,DBLP:conf/iccvw/YanZXZWS19} proposed to replace KL by the Euclidean distance, i.e. mean squared error (MSE) for feature-based distillation. MSE is widely adopted in the recent SOTA approaches \cite{ReFO} and is given by:
\begin{equation}
\label{eq:mse}
\vspace{-2mm}
    \mathcal{L}_{mse}= \frac{1}{N}  \sum\limits_{i \in n}  \Vert f_i^s - f_i^t \Vert_2^2,
\end{equation}
where $N$ is the batch size. 
Feature distillation with MSE proves to improve $S$ model verification accuracies \cite{ReFO}. However, this approach faces two main challenges. First, it assumes that the compact student model can learn to imitate the teacher embedding spaces using a shallow network architecture.   However, achieving this goal is challenging due to the large gap, in terms of network architecture size between the compact $S$ and the large $T$, as well as the limited learning capability of the shallow $S$ network \cite{ReFO,DBLP:conf/iccvw/YanZXZWS19}.
Several previous works proposed to address this challenge using, for example, multiple phases of student and teacher training \cite{DBLP:conf/iccvw/YanZXZWS19,ReFO} or by using teacher assistants \cite{DBLP:conf/aaai/MirzadehFLLMG20}. which is computationally expensive, especially during the training phase.
Second, as feature distillation with MSE aims at imitating the exact teacher embedding space \cite{ReFO,DBLP:conf/iccvw/YanZXZWS19} (which is challenging due to the capacity gap between $S$ and $T$), it prevents the student model from establishing its own embedding space or explicitly learning the proximity between samples and their class centers and negative class centers.
We therefore propose in this work an adaptive feature-based KD approach that addresses these challenges by 1) Instead of imitating the exact teacher embedding space, we train the student to learn to push samples to be relatively close to their class centers and far from other class centers, and 2)  we further propose an adaptive approach that guided the student model to learn less complex knowledge at an early stage of training and more complex knowledge at a later stage of training.

\vspace{-1mm}
\subsection{Feature distillation through classification loss}
\paragraph{Preliminary on classification loss function:}
The Softmax loss and its variants, i.e. margin penalty softmax losses, are commonly used loss functions to train FR models \cite{deng2019arcface,DBLP:conf/cvpr/LiuWYLRS17,DBLP:conf/cvpr/WangWZJGZL018} that implicitly learn to produce discriminative feature representations through learning a multiclass classification. The Softmax loss refers to applying cross-entropy between the output of the softmax activation function and ground-truth labels. Margin penalty softmax losses extend the softmax loss by introducing a margin penalty on the cosine angle (or its cosine) of the feature representation of sample $i$ and its class center.
This aims at pushing the decision boundary of the softmax, in turn enhancing intra-class compactness and inter-class discrepancy. In this work, we utilize two versions of the margin penalty softmax loss, ArcFace \cite{deng2019arcface} and CosFace \cite{DBLP:conf/cvpr/WangWZJGZL018}.
It should be noted that our proposed approach can be deployed using any margin penalty softmax loss. We opt for ArcFace and CosFace, as examples, as they are widely used classification losses in the FR literature \cite{deng2019arcface,DBLP:conf/cvpr/WangWZJGZL018}.
Formally, the Angular Margin Penalty-based Losses (AML), ArcFace and CosFace, are given by:
\begin{equation}
\label{eq:arc}
\vspace{-2mm}
    \mathcal{L}_{AML}=\frac{1}{N}  \sum\limits_{i \in N} - log \frac{e^{s (cos(\theta_{y_i}+m1) - m2)}}{ e^{s(cos(\theta_{y_i}+m1) - m2)} +\sum\limits_{j=1 , j \ne y_i}^{c}  e^{s ( cos(\theta_{j}))}},
\end{equation}
where $N$ is the batch size, $y_i$ is the class label of sample $i$ (in range $[1,c]$, where $c$ is the number of classes), $\theta_{y_i}$ is the angle between the features $f_{i}$ and the $y_i$-th class center $w_{y_i}$. {$f_{i} \in \mathbb{R}^d$}  is the deep feature embedding of the last fully connected layer of size $d$. $w_{y_i} \in \mathbb{R}^d$ is the $y_i$-th column of weights $W \in \mathbb{R}^{d \times c}$ of the classification layer.
$\theta_{y_i}$ is defined as ${f_iw^T_{y_i}}=\Vert f_i \Vert \Vert w_{y_i} \Vert cos(\theta_{y_i})$ \cite{DBLP:conf/cvpr/LiuWYLRS17}. The weights and the feature norms are fixed to $\Vert w_{y_i} \Vert=1$ and $ \Vert f_i \Vert =1$, respectively, using $l_2$ normalization, as defined in \cite{DBLP:conf/cvpr/LiuWYLRS17,DBLP:conf/cvpr/WangWZJGZL018}. The decision boundary, in this case, depends on the cosine of the angle between $f_{i}$ and $w_{y_i}$.
In ArcFace \cite{deng2019arcface}, $m1>0$ and $m2=0$, where $m1$ is an additive angular margin penalty, aiming at enhancing the intra-class compactness and inter-class discrepancy. 
In CosFace \cite{DBLP:conf/cvpr/WangWZJGZL018}, $m1=0$ and $m2>0$ is additive cosine margin penalty.
Lastly, $s$ is the scaling parameter \cite{DBLP:conf/cvpr/WangWZJGZL018}.
The decision boundary of AML is $cos(\theta_{y_i}+m1)-m2 -  cos(\theta_j)=0$. The model, in this case, learns to optimize $f_i$ and $w_{j}$ as $cos(\theta_j) \leq cos(\theta_{y_i}+m1)-m2$, i.e., pushing the sample $f_i$ to be closer to its class center than other negative class centers. 
In the backpropagation phase, the weights of both $f_{i}$ and $w_{j}$ are updated incrementally after each iteration using the Gradient Decent optimization algorithm \cite{DBLP:journals/corr/Ruder16}.
In the following sections,  $\mathcal{L}_{AML}$ will be denoted as $\mathcal{L}_{Arc}$ and $\mathcal{L}_{Cos}$ for ArcFace and CosFace, respectively.
\begin{figure}[!]
\vspace{-3mm}
     \centering
     \begin{subfigure}[b]{0.32\linewidth}
         \centering
         \includegraphics[width=\linewidth]{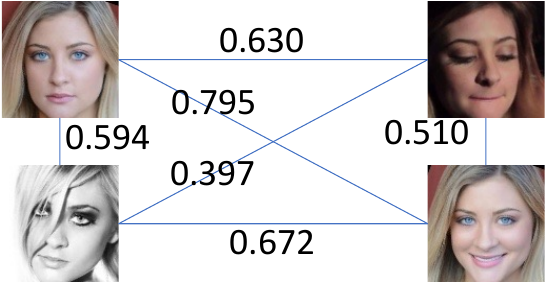}
         \caption{Sample-Sample
         }
         \label{fig:sample_sample}
     \end{subfigure}
     \begin{subfigure}[b]{0.32\linewidth}
         \centering
         \includegraphics[width=\linewidth]{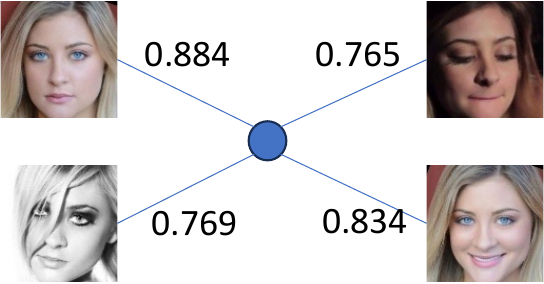}

         \caption{Sample-Center
         }
         \label{fig:sample_center}
     \end{subfigure}
     \begin{subfigure}[b]{0.32\linewidth}
         \centering
         \includegraphics[width=\linewidth]{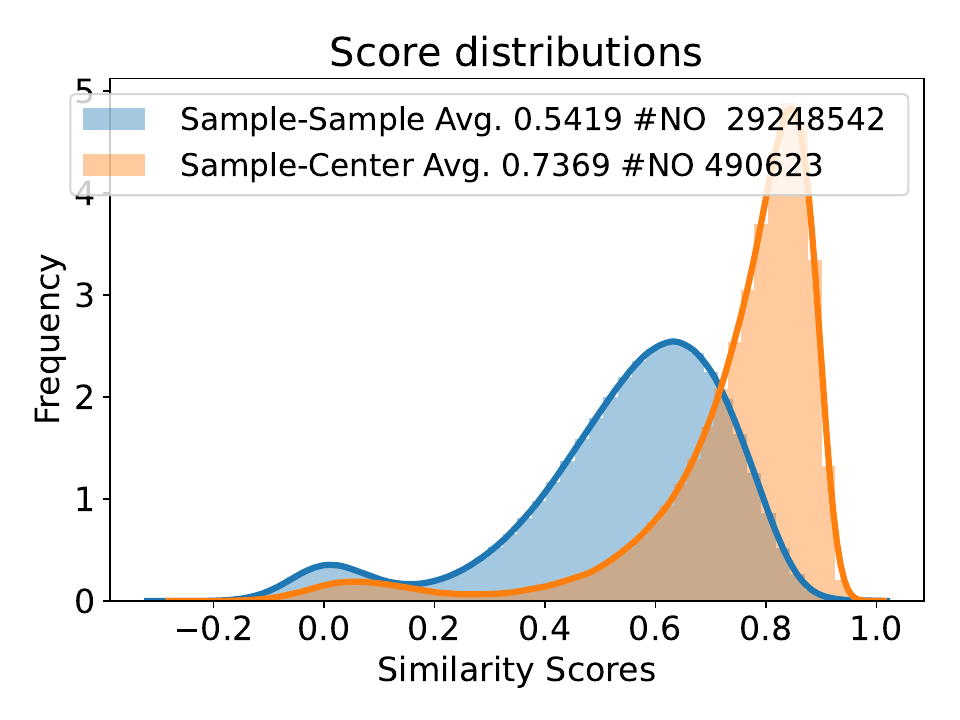}

         \caption{Score distributions
         }
         \label{fig:scores_distribution}
     \end{subfigure}
     \vspace{-2mm}
      \caption{ Sample-sample and Sample-center similarity score distributions. Fig. \ref{fig:sample_sample} and \ref{fig:sample_center} presented an example of matching samples with others (of the same identity) and samples with their class center.  Fig. \ref{fig:scores_distribution} matching samples with samples (of the same identity) achieved lower similarity scores than matching samples with their class centers.
     }
        \label{fig:sample_center_score}
        \vspace{-6mm}
\end{figure}

\textbf{AMLDistill: Additive margin penalty distillation:}
The class centers $w_{j}^t$ learned by the teacher model contain more identity-representative information than $f_i^t$ of one sample.  Fig. \ref{fig:scores_distribution} presents two similarity score distributions obtained by comparing samples with all other samples of each identity (blue plot) and samples with their class centers (orange plot). It can be noticed that samples can match with their class center with a higher similarity score than matching sample-to-sample of the same identity, indicating that class centers better represent the identity classes than any single sample. For the visualization, the similarity scores are calculated using the cosine similarity between embeddings\footnote{We utilized ResNet100 \cite{ResNet} trained using ArcFace loss \cite{deng2019arcface} on MS1MV2 dataset \cite{guo2016ms,deng2019arcface} to obtain the embedding} from CASIA-WebFace (0.5M images). 
Thus, we train $S$, similarly to \cite{MarginKD}, to optimize the angular distance between $f_i^s$ (from $S$) and $w_{j}^t$ (from $T$).  
Specifically, we propose to train the student model to push $f_i^s$ in a way that is closer to its class center $w_{yi}^t$ and farther from other negative class centers $w_{j}^t$. 
Specifically, we set $w_{j}^s=w_{j}^t$. The modified loss function is given by:
\begin{equation}
\label{eq:arcdistill}
\vspace{-2mm}
\mathcal{L}_{AMLDistill}=\frac{1}{N}  \sum\limits_{i \in N} - log \frac{e^{s (cos(\theta_{y_i}^t+m1)-m2)}}{ e^{s(cos(\theta_{y_i}^t+m1) -m2)} +\sum\limits_{j=1 , j \ne y_i}^{C}  e^{s ( cos(\theta_{j}^t))}},
\end{equation}
where $\theta_{y_i}^t$ is the angle between the features $f_i^s$ and the $y_i$-th class center $w_{y_i}^t$ learned by the teacher model.
$cos(\theta_{y_i}^t)$, in this case, is defined as follow:
\begin{equation}
\vspace{-2mm}
  cos(\theta^t_{y_i})=  {f_i^s(w^{t}_{y_i}})^T,
\end{equation} 
where $f_i^s$ is the feature representation learned by the student and $w_{yi}^t$ is its class.

In the backpropagation phase, the weights $w_{j}^t$ are fixed and only $f_{i}^s$ are updated after each epoch using the Gradient Decent optimization algorithm \cite{DBLP:journals/corr/Ruder16}.
Unlike previous KD approaches \cite{ShrinkTeaNet,EKD,RKD,FitNet,CCKD,ReFO}, AMLDistill (Eq. \ref{eq:arcdistill} ) requires access only to the precomputed weight of the last layer of $T$ (i.e., $w^s_{j}=w^t_{j}$) and does not require passing the training samples into the teacher model.
Our models are denoted as ArcDistill, when the utilized margin penalty is $m1$ (ArcFace) and as CosDistill, when the utilized margin penalty is $m2$ (CosFace).

\begin{figure*}[t!]
  \centering
  \begin{subfigure}[b]{0.28\linewidth}
         \centering
   \includegraphics[width=\linewidth]{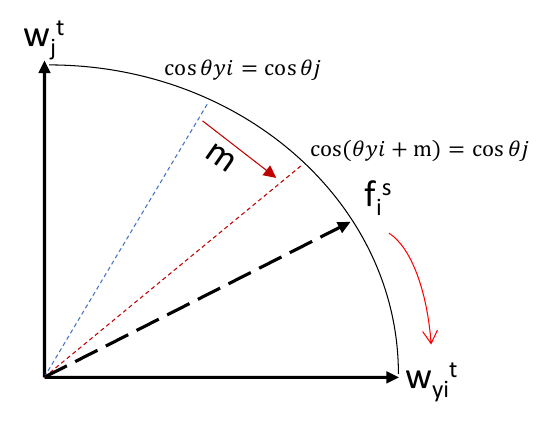}
         \caption{ArcDistill}
         \label{fig:arcdistill}
     \end{subfigure}
     \begin{subfigure}[b]{0.28\linewidth}
         \centering
   \includegraphics[width=\linewidth]{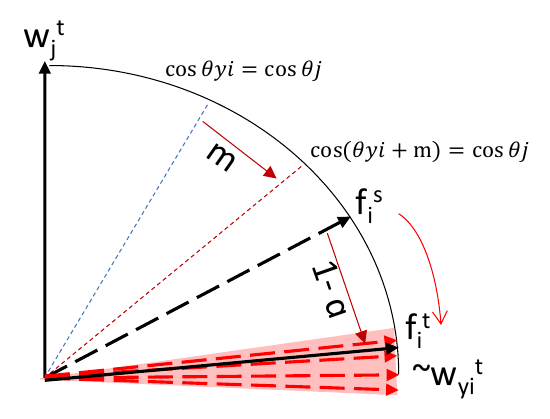}
         \caption{AdaDistill (Early-stage)}
         \label{fig:adadistill_earlystage}
     \end{subfigure}
     \begin{subfigure}[b]{0.28\linewidth}
         \centering
   \includegraphics[width=\linewidth]{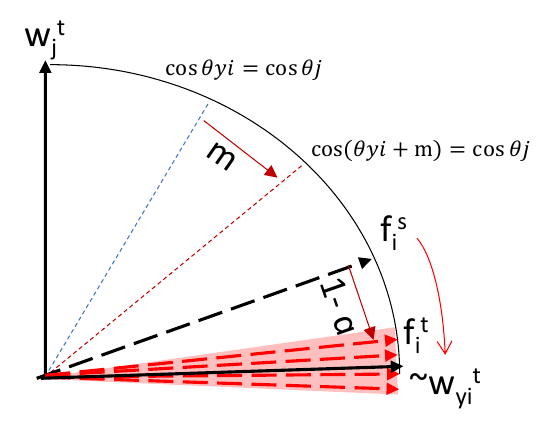}
         \caption{AdaDistill (Later-stage)}
         \label{fig:adadistill_latterstage}
     \end{subfigure}
     \vspace{-3mm}
   \caption{Different distillation approaches using fixed class center (Fig.  \ref{fig:arcdistill}) and adaptive class center (Figs. \ref{fig:adadistill_earlystage} and \ref{fig:adadistill_latterstage}). 
   In ArcDistill (Figrue \ref{fig:arcdistill}),  $f_i^s$ is pushed with penalty margin to fixed class center $w_{yi}^t$. 
   Figs. \ref{fig:adadistill_earlystage} and \ref{fig:adadistill_latterstage} illustrate the adaptive estimation of class center $w_{yi}^t$ using exponential moving average (Eq. \ref{eq:w}). 
   At an early stage of training (Fig. \ref{fig:adadistill_earlystage}), the $\alpha$ value (positive cosine similarity between $f_i^s$ and $f_i^t$) is small and the class center is close to $f_i^t$. 
   As training progresses (Fig. \ref{fig:adadistill_latterstage}), $\alpha$ increases and the new class center is estimated to be close to the average of all $f_i^t$ of class $y_i$.  
   }
   \label{fig:boundaries}
   \vspace{-5mm}
\end{figure*}

\vspace{-3mm}
\subsection{AdaDistill: Adaptive AMLDistill}
\begin{figure}[t!]
\vspace{-3mm}
     \centering
     \begin{subfigure}[b]{0.47\linewidth}
         \centering
    \includegraphics[width=\textwidth]{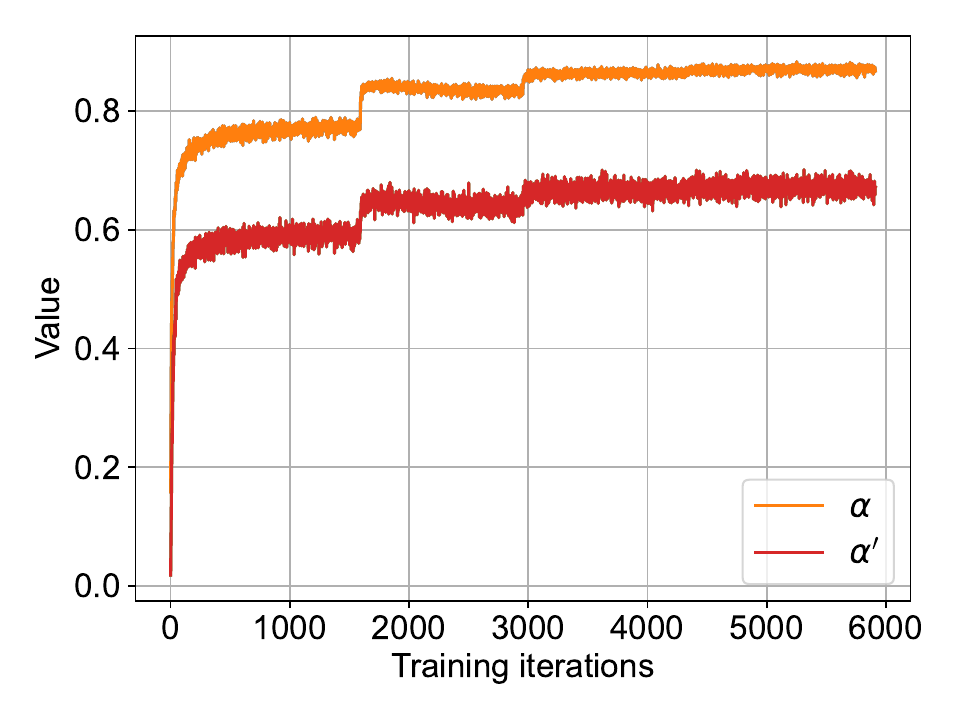}
    \caption{
         }
         \label{fig:alpha_values}
     \end{subfigure}
     \begin{subfigure}[b]{0.47\linewidth}
         \centering
         \includegraphics[width=\linewidth]{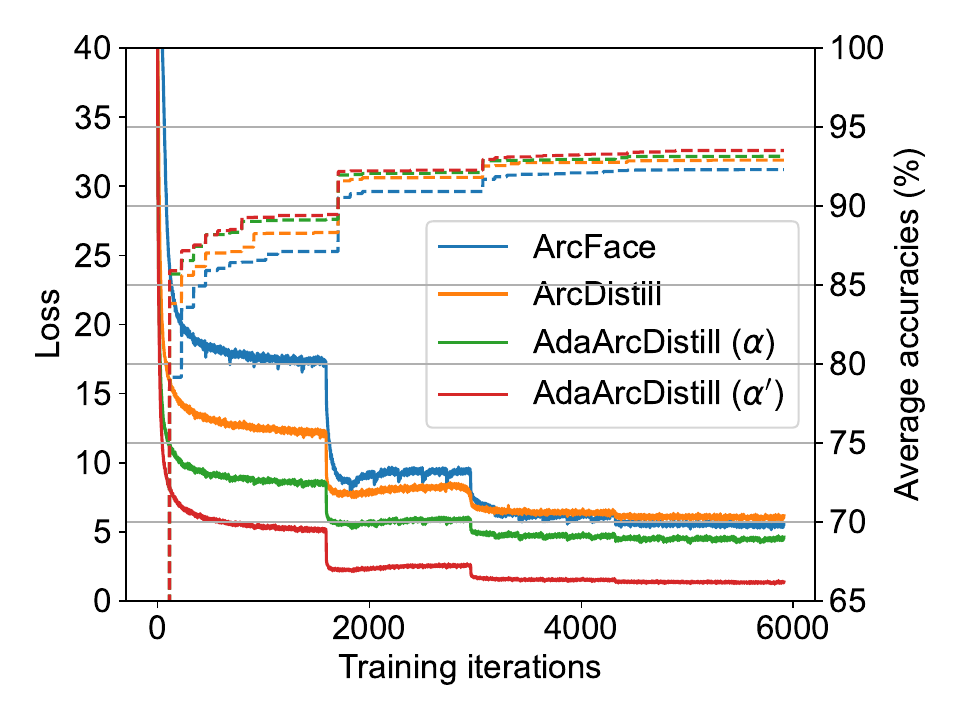}
    \vspace{-3mm}
    \caption{
    }
          \label{fig:loss}
     \end{subfigure}
     \vspace{-4mm}
      \caption{ 
      Fig. \ref{fig:alpha_values} (Left): Average 
    $\alpha$ and $\alpha'$ values over the training iterations as defined in Eq.\ref{eq:alpha} and \ref{eq:w_alpha}, respectively.
    Fig. \ref{fig:loss} (Right):     Loss values and average verification accuracies over training iterations of students trained with different loss functions. The losses are in solid lines and the average accuracies are in dashed.
    The average accuracies are calculated on five benchmarks, LFW \cite{LFWTech}, CFP-FP \cite{cfp-fp}, AgeDB-30 \cite{agedb}, CA-LFW \cite{CALFW} and CP-LFW \cite{CPLFWTech}, described in Section \ref{sec:dataset}.  
    AdaDistill ($\alpha$) and AdaDistill ($\alpha'$) facilitated better convergence and achieved higher accuracies in comparison to the ArcDistill and standalone ArcFace.
     }
        \label{fig:meregend}
        \vspace{-6mm}
\end{figure}

\vspace{-3mm}
Given that $w_{yi}^t$ captures more complex and identity-representative information than $f_{i}^t$ and considering the substantial discrepancy between $T$ and $S$ model, $S$ is not fully capable of mimicking the distance between $f_i^s$ and the teacher-centers  $w_{yi}^t$. This can be observed from Fig. \ref{fig:loss}, which presents the training losses over the training iterations of various models trained with different loss functions. It can be clearly noticed that the model trained with ArcDistill (orange plot), i.e., $S$ aimed at minimizing the distance between $f_i^s$ and the teacher-centers  $w_{yi}^t$, suffers from low convergence in comparison to AdaArcDistill, which we introduce in the remainder of this section. To address the aforementioned challenges, we propose an approach based on Exponential Moving Average (EMA) to refine $w_{yi}^t$ over the model training stage. 
Specifically, given that the average of all $f_i^t$ approximately corresponds to the $i-th$ class center, we propose to estimate $w_{yi}$ using $f_i^t$ in different training stages. 
Let $\theta^{easy}$ and $\theta^{hard}$ be the angle between $f_i^s$ and $f_i^t$ and $f_i^s$ and $w_{yi}$, respectively.
The key idea is that the student at an early stage of training can more easily learn $f_i^s$ as $cos(\theta_j^{easy}) \leq cos(\theta_{y_i}^{easy}+m)$ than $cos(\theta_j^{hard}) \leq cos(\theta_{y_i}^{hard}+m)$. 
As the training progresses, $S$ will be more capable of learning complex patterns from the identity center.   
Formally, $w_{yi}$ is updated after each $k-th$ training iteration as:
\begin{equation}
\label{eq:w}
\vspace{-2mm}
    w_{yi}^{(k)}= \alpha  w_{yi}^{(k-1)} + (1 - \alpha)  (f_{i}^{t(k)})^T,
\end{equation}
where $k$ is the current training steps. $T$ is the transpose function.
$\alpha \in [0,1]$ is the momentum parameter and set to the positive cosine similarity  between $f_i^s$ and $f_i^t$ from the $k-th$ batch, given by:
\begin{equation}
\label{eq:alpha}
\vspace{-2mm}
    \alpha= \lfloor Cos(f_i,f_i^t) \rceil^1_0,
\end{equation}
where $\lfloor.\rceil$ refers to clipping the value between 0 and 1. 
Based on the refined $w$, the student model is trained as in Eq. \ref{eq:arcdistill}. At an early stage of training, $\alpha \approx 0$ and thus, $cos(\theta_{y_i}^t) = {f_i^s(w_{yi})^{T}}, w_{yi} \approx f_i^t$.  The student, in this case, learns to minimize the distance between $f_i^s$ and its counterpart $f_i^t$. This early training stage is referred to as an easy stage, where $S$ learns less complex information.
As seen in Fig. \ref{fig:alpha_values}, the $\alpha$ value is increased over the training iterations. At a later stage of training $\alpha \approx 1$ and thus, $cos(\theta_{y_i}^t) = {f_i^s  (w_{yi})^{T}}, w_{yi} \approx Avg(f_{yi}^t)$, i.e, $w^{t}_{y_i}$ is the average embedding of $f^t$ belong to class $y_i$. In this later stage of training, $S$ learns complex information about identity centers.
This adaptive distillation function at the early stage and the late stage is simplified visually, along with the function when the class center is fixed, in Fig. \ref{fig:boundaries}. 

\textbf{Importance of hard mining:}
The $\alpha$, defined in Eq. \ref{eq:alpha}, represents the ability of the student model to mimic the teacher feature space, and thus, the student fully controls $\alpha$. In this case, there is no consideration of the importance of the hard samples as induced by the teacher model. For that, we propose to weight the $\alpha$ based on the importance of the hard training sample. Specifically,
we estimate the difficulty of each training sample in the current batch based on the positive cosine similarity between feature representation $f_i^t$ of sample $i$ (obtained from the teacher) and its estimated class center $w_{yi}$. The weighted $\alpha'$ is given by:
\begin{equation}
\label{eq:w_alpha}
\vspace{-2mm}
    \alpha'= \lfloor Cos(f_i^s,f_i^t) \times  Cos((w_{yi}^{(k-1)})^T,f_i^t) \rceil^1_0.
\end{equation}
In each training iteration, when the value of the weighting term $\smash{Cos((w_{yi}^{(k-1)})^T,f_i^t)}$ is small, i.e, hard sample, $\alpha$ (Eq. \ref{eq:w}) is small and thus $\smash{w_{yi}^{(k)}}$ is refined toward $f_i^t$ of hard sample. 
Conversely, when $\smash{Cos((w_{yi}^{(k-1)})^T,f_i^t)}$ is large, the center $ w_{yi}^{(k)}$ is refined toward the average embeddings. The average value of $\alpha'$ over training iteration is shown in Fig. \ref{fig:alpha_values}.
Our AdaDistill is used to train $S$ with loss defined  \ref{eq:arcdistill} using refined $w_{yi}$ (Eq \ref{eq:w}) and weighted $\alpha'$ (Eq .\ref{eq:w_alpha}).
All our models are trained with only the AdaArcDistill, and thus, it mitigates the need for optimizing different learning objectives or tuning loss weighting terms. Our models are noted as AdaArcDistill when the utilized margin penalty is $m1$ (ArcFace) and as AdaCosDistill when the utilized margin penalty is $m2$ (CosFace).

\vspace{-5mm}
\section{Experimental settings}
\vspace{-3mm}
\subsection{Datasets}
\label{sec:dataset}
\vspace{-2mm}
\textbf{Training:} We use MS1MV2 \cite{guo2016ms,deng2019arcface}  to train our proposed models for fair comparisons with SOTA KD approaches \cite{EKD,ReFO}.  MS1MV2 is a refined version of  MS-Celeb-1M \cite{guo2016ms} by \cite{deng2019arcface}, containing 5.8M images of 85K identities.  
In our ablation study, we use also Casia-WebFace \cite{CASIAWbFace} and IDiff-Face \cite{Boutros_2023_ICCV} in evaluating training the student using a different dataset than the one used to train the teacher. 
Casia-WebFace \cite{CASIAWbFace} contains 494,414 images of 10,575 different identities. IDiff-Face \cite{Boutros_2023_ICCV} is a synthetic dataset generated by a conditional latent diffusion model and it contains 500k images of 10k identities.
Note that all results in this paper are reported using MS1MV2 as a training dataset unless explicitly noted. 

\textbf{Testing:} 
The presented models are evaluated on 10 diverse benchmarks:  Labeled Faces in the Wild (LFW) \cite{LFWTech}, AgeDB-30 \cite{agedb}, Celebrities in Frontal-Profile in the Wild (CFP-FP) \cite{cfp-fp}, Cross-age LFW (CA-LFW) \cite{CALFW}, Cross-Pose LFW (CP-LFW)  \cite{CPLFWTech}, ICCV21-MFR \cite{DBLP:conf/iccvw/DengGAZZ21}, IARPA Janus Benchmark–C \cite{ijbc} (IJB-C) and Benchmark–B \cite{ijbb}, MegaFace \cite{DBLP:conf/cvpr/Kemelmacher-Shlizerman16}, and the refined MegaFace dataset (MegaFace (R)) \cite{deng2019arcface,DBLP:conf/cvpr/Kemelmacher-Shlizerman16}.  

\textbf{Data preprocessing:} Following previous works \cite{ReFO,EKD,deng2019arcface}, all facial images are aligned and cropped to $112 \times 112$, as described in \cite{deng2019arcface}, using five landmark points extracted by the Multi-task Cascaded Convolutional Networks (MTCNN) \cite{MTCNN}. Also, images are normalized to have pixel values between -1 and 1 before passing them into the student and teacher networks.

\vspace{-4mm}
\subsection{Evaluation metric}
\label{sec:metric}
\vspace{-2mm}
The verification performances on LFW \cite{LFWTech}, AgeDB-30 \cite{agedb}, CA-LFW \cite{CALFW}, CP-LFW \cite{CPLFWTech}, and CFP-FP \cite{cfp-fp} are reported in the form of verification accuracy (\%), following their evaluation protocols.
We followed the 1:1 mixed verification protocol of IJB-B and IJB-C and report the result as true acceptance rates (TAR) at false acceptance rates (FAR) of 1e-4 and 1e-5 \cite{ijbb,ijbc}. The ICCV21-MFR \cite{DBLP:conf/iccvw/DengGAZZ21} contains three test sets: Mask, Children, and Multi-racial (MR-all). Following the ICCV21-MFR evaluation metrics \cite{DBLP:conf/iccvw/DengGAZZ21}, we report the results of the Mask and Children test sets as 
TAR at FAR=1e-4 and MF-all as TAR at FAR=1e-6.
The MegaFace and MegaFace(R) benchmarks report the recognition performance as Rank-1 identification rate and as TAR at FAR=1e–6 verification accuracy.
\vspace{-4mm}
\subsection{Implementation details}
\vspace{-2mm}
\textbf{Network architectures:} We employ the pretrained ResNet50 (43.59 m parameters and 13.64 GFLOPs) \cite{ResNet,deng2019arcface} as the teacher. The teacher is trained on MS1MV2 with ArcFace using default training settings (scale parameter $s$ is 64 and the margin $m$ is 0.5) \cite{deng2019arcface}.
The weights of the teacher model in all experiments are frozen.
The feature embedding size is 512. 
Following previous SOTA approaches, we use  MobileFaceNet \cite{MFN} (1.19m parameters and 0.45 GFLOPs), noted as MFN,  as the student.
Our choice for the teacher (ResNet50) and the student (MFN) is to be aligned and to provide a fair comparison with SOTA methods \cite{ReFO,MarginKD,EKD}.
Additionally, we evaluate our AdaDistill using different pretrained teachers' architectures, ResNet18 (24.02m parameters and 5.23 GFLOPs) \cite{ResNet,deng2019arcface},  ResNet100 (65.15m parameters and 24.21 GFLOPs) \cite{ResNet,deng2019arcface} and TransFace-B (124.47m parameters and 21.92 GFLOPs) \cite{TransFace}. 

\textbf{Student training setups}
The presented models are implemented using Pytorch \cite{NEURIPS2019_9015}. We followed the default setting of ArcFace \cite{deng2019arcface} and CosFace \cite{DBLP:conf/cvpr/WangWZJGZL018} and set the scale parameter $s$ to 64 and margin $m1$ to 0.5 in ArcFace and  $m2$ to 0.35 in CosFace(Eq. \ref{eq:arcdistill}). The selection of the margin values is validated in our ablation study.
The batch size in all experiments is set to 512 and the models are trained with a Stochastic Gradient Descent (SGD) optimizer with an initial learning rate of 1e-1. 
The momentum is set to 0.9 and the weight decays to 5e-4. The learning rate is divided by 10 at 80k, 140k, 210k, and 280k training iterations. During the student training, only random horizontal flipping with a probability of 0.5 is used as data augmentation. The models are trained on 4 Nvidia GeForce RTX 6000 GPUs. Under the same settings, training standalone MFN requires an average of 0.087 seconds to process (forward and backward propagation) a batch of 512 samples.
Training MFN with ResNet50 using vanilla KD (Eq. \ref{eq:mse}) requires an average of 0.207 seconds for a single batch. This time complexity is reduced to 0.134 seconds using ArcDistill. Our AdaArcDistill ($\alpha$) and AdaArcDistill ($\alpha'$) add a slight burden on training complexity, compared to vanilla KD, with 0.243 and 0.247 seconds, respectively to process a single batch. 


\begin{table*}[t!]
\centering
\resizebox{0.95\textwidth}{!}{%
\begin{tabular}{llllllll|ll}
\hline
\multirow{2}{*}{Ablation}                      & \multirow{2}{*}{Method}                   & \multirow{2}{*}{LFW} & \multirow{2}{*}{CFP-FP} & \multirow{2}{*}{AgeDB-30} & \multirow{2}{*}{CA-LFW} & \multirow{2}{*}{CP-LFW} & \multirow{2}{*}{Avg.} & \multicolumn{2}{l}{IJBC}        \\
                                               &                                           &                      &                         &                          &                         &                         &                       & 1e-4           & 1e-5           \\ \hline
                                               & ResNet50 (Teacher)                        & 99.80                & 97.63                   & 97.92                    & 96.10                   & 92.43                   & 96.77                 & 96.05          & 93.96          \\
                                               & MFN (Student)                             & 99.52                           & 91.66                           & 95.82                           & 95.12                           & 87.93                           & 94.01  & 89.13                     & 81.65         \\ \hline
\multirow{4}{*}{1) KD method}                       & MFN + Vanilla KD (Eq. \ref{eq:mse})            & 99.57                & 93.77                   & 96.97                    & 95.70                   & 89.01                   & 95.00                 & 91.29          & 79.79          \\
                                               & MFR + ArcDistill     (Eq. \ref{eq:arcdistill})                      & 99.63                & 93.83                   & 96.82                    & 95.48                   & 89.20                   & 94.99                 & 91.70          & 84.57          \\
                                               & MFN + AdaArcDistill ($\alpha$)   (Eq. \ref{eq:arcdistill}, \ref{eq:w} and \ref{eq:alpha})          & 99.60                & 94.00                   & 96.88                    & 95.65                   & 89.63                   & 95.15                 & 92.56          & 88.23          \\
                                               & MFN + AdaArcDistill ($\alpha'$)  (Eq. \ref{eq:arcdistill}, \ref{eq:w} and \ref{eq:w_alpha})         & 99.60                & 95.04                   & 96.78                    & 95.58                   & 90.08                   & \textbf{95.42}        & \textbf{92.97} & \textbf{89.13} \\ \hline
\multirow{3}{*}{2) Margin Penalty} & MFN + AdaArcDistill ($\alpha'$), m =0.40  & 99.53                & 95.21                   & 96.82                    & 95.54                   & 90.02                   & 95.42                 & 92.98          & 88.87          \\
                                   \multirow{3}{*}{type and value}             & MFN + AdaArcDistill ($\alpha'$),  m =0.45 & 99.58                & 95.24                   & 96.78                    & 95.55                   & 90.02                   & \textbf{95.47}        & \textbf{93.27} & \textbf{89.32} \\
                                               & MFN + AdaArcDistill ($\alpha'$),  m =0.50 & 99.60                & 95.04                   & 96.78                    & 95.58                   & 90.08                   & 95.42                 & 92.97          & 89.13          \\ \cline{2-10} 
                                               & MFN + AdaCosDistill ($\alpha'$), m=30     & 99.50                & 94.96                   & 95.95                    & 95.20                   & 90.32                   & 95.19                 & 92.92          & \textbf{89.26}          \\
                                               & MFN + AdaCosDistill ($\alpha'$), m=35     & 99.50                & 95.07                   & 96.60                    & 95.33                   & 90.02                   & \textbf{95.30 }                & \textbf{93.05 }         & 89.21          \\
                                               & MFN + AdaCosDistill ($\alpha'$), m=40     & 99.60                & 93.98                   & 96.57                    & 95,45                   & 89.35                   & 94.99                 & 93.12          & 89.09          \\ \hline
\multirow{6}{*}{3) Teacher architecture } & 

ResNet18 (Teacher)      & 99.67                & 94.47                   & 97.13                    & 95.70                   & 89.73                   & 95,34                 & 93.99       & 91.14      \\
& MFN + AdaArcDistill($\alpha'$), m=45    & 99.60                & 94.30                   & 96.43                    & 95.35                   & 89.13                   & 94,96                 & 92.27       & 88.10      \\ \cline{2-10} 
& ResNet50 (Teacher)      & 99.80                & 97.63                   & 97.92                    & 96.10                   & 92.43                   & 96.77                 & 96.05       & 93.96      \\
& MFN + AdaArcDistill($\alpha'$), m=45     & 99.58                & 95.24                   & 96.78                    & 95.55                   & 90.02                   & 95.47                 & 93.27       & 89.32      \\ \cline{2-10} 
& ResNet100 (Teacher)     & 99.83                & 98.40                   & 98.33                    & 96.13                   & 93.22                   & 97,19                 & 96.39       & 94.58      \\
& MFN + AdaArcDistill($\alpha'$), m=45     & 99.63                & 94.26                   & 96.97                    & 95.60                   & 89.28                   & 95,15                 & 92.54       & 87.48      \\  \cline{2-10}
& TransFace-B  (Teacher)         & 99.85 & 99.17  & 98.53          &  96.20      &  92.92    & 97.33   & 96.55        & 94.15        \\ 
& MF+  AdaArcDistill($\alpha'$), m=45 & 99.57 & 94.23  & 96.53    & 95.47  & 89.87  &    95.13   & 92.85        & 87.98   \\
\hline 
\end{tabular}
}
\caption{Ablation studies on different KD methods, margin parameters selection, and different teacher architecture. The verification accuracies are reported in (\%) on five small benchmarks, LFW, CFP-FP, AgeDB-30, CA-LFW, CP-LFW, and as TAR@FAR of 1e-4 and 1e-5 on large-scale IJB-C benchmark.
The results are grouped based on the targeted ablation study. 
The first two rows present the results of the student and the teacher (used in the 1st and 2nd studies ), respectively.}
\label{tab:ablation_study}
\vspace{-7mm}
\end{table*}

\vspace{-6mm}
\section{Results}
\vspace{-3mm}
In this section, we present rigorous ablation studies to analyze the characteristics of  AdaDistill as well as comprehensive comparisons against competing SOTA methods. Additional results using different student architecture are presented in the Supplementary material.
\vspace{-3mm}
\subsection{Ablation study}
\vspace{-3mm}
Table \ref{tab:ablation_study} presents the verification performances on six benchmarks. We use the average accuracy on the five small benchmarks (LFW, CFP-FP, AgeDB-30, CAL-LFW, and CP-LFW) and the verification accuracy on the large-scale IJB-C as the main criterion for comparing different methods and parameters selection in our ablation and sensitivity studies. The results are reported in (\%) using the evaluation metrics described in Section \ref{sec:metric}.
The first two rows present the performance of the teacher (ResNet50) and the student (MFN) models, respectively.

\textbf{Impact of adaptive class centers and importance of hard samples:}
We first conducted several experiments on MFN using different KD methods, including 1) Vanilla KD (Eq.\ref{eq:mse}), 2) ArcDistill (Eq.\ref{eq:arcdistill}), AdaArcDistill (Eq.\ref{eq:arcdistill}, \ref{eq:w} and \ref{eq:alpha}) and AdaArcDistill (Equations \ref{eq:arcdistill}, \ref{eq:w} and \ref{eq:w_alpha}). 
All distillation methods improved the verification accuracies of the student model. This observation is demonstrated by comparing MFN (second row in Table \ref{tab:ablation_study}) with the first group of the ablation study (KD method) in Table \ref{tab:ablation_study}.
The model trained with the adaptive class centers outperformed the one trained with fixed class centers.
It can be also noticed that the largest improvement is achieved using our AdaArcDistill($\alpha'$), emphasizing the benefit of the adaptive class centers and the importance of hard samples. 
These enhancements in the student model performance are also noticed in the model convergence in Fig. \ref{fig:loss}, where we plotted the loss values over the training iteration of student models trained with different loss functions. Our AdaArcDistill($\alpha'$) facilitated better convergence in comparison to ArcDistill and the standalone student (ArcFace).

\textbf{Study on margin penalty:}
We evaluated the impact of the type and the value of the margin penalty by utilizing two additive margin penalty losses, additive angular margin penalty (AdaArcDistill Eq. \ref{eq:arcdistill}) and additive cosine margin penalty (AdaCosDistill Eq. \ref{eq:arcdistill}).  For each of the considered losses, we evaluated different penalty values, following their respective works \cite{deng2019arcface,DBLP:conf/cvpr/WangWZJGZL018}. As shown in Table \ref{tab:ablation_study}, ArcFace with $m$ of 0.45 achieved the best overall performance. For CosFace, the best performance is achieved using $m$ of 0.35. 
It should be noted that the best performance in the ArcFace \cite{deng2019arcface} and in the CosFace \cite{DBLP:conf/cvpr/WangWZJGZL018} papers, when used as standalone FR training, are reported using $m=0.5$ and $m=0.35$, respectively. 
It can be also observed from Table \ref{tab:ablation_study} that both AdaArcDistill ($m=0.45$) and AdaCosDistill ($m=0.35$) achieved very competitive results. This indicates the effectiveness of our AdaDistill on different losses. It should be also noted that AdaDistill does not require tuning any additional hyper-parameters.


\textbf{Different teacher architectures:}
The last group of results in Table \ref{tab:ablation_study} presents the verification accuracies of MFN models trained with our AdaArcDistill using three different teacher architectures, TransFace-B, ResNet100, ResNet50, and ResNet18.
As expected, KD from all teacher models improved the MFN performances. 
The teachers ResNet100 and TransFace-B, as expected, achieved the highest verification accuracies compared to ResNet50 and ResNet18. However, it can be noticed that ResNet50, as a teacher, leads to better distillation performances, in comparison to the weaker teacher (ResNet18) and stronger teacher (ResNet100 and TransFace-B). 
These results indicate the importance of choosing the right teacher model for KD and the applicability of the AdaDistill concept from teachers with different gaps to the student model.

\begin{table*}[t!]
\centering
\resizebox{0.95\linewidth}{!}{%
\begin{tabular}{llllllll|ll}
\hline
\multirow{2}{*}{Method} & \multirow{2}{*}{Training Dataset} & \multirow{2}{*}{LFW} & \multirow{2}{*}{CFP-FP} & \multirow{2}{*}{AgeDB-30} & \multirow{2}{*}{CA-LFW} & \multirow{2}{*}{CP-LFW} & \multirow{2}{*}{Avg.} & \multicolumn{2}{l}{IJBC} \\
                        &                                   &                      &                         &                          &                         &                         &                       & 1e-4        & 1e-5       \\ \hline
ResNet50 (Teacher)      & MS1MV2  \cite{guo2016ms,deng2019arcface}                          & 99.80                & 97.63                   & 97.92                    & 96.10                   & 92.43                   & 96.77                 & 96.05       & 93.96      \\ \hline
MFN (Student)           & \multirow{2}{*}{IDiff-Face \cite{Boutros_2023_ICCV}}       &  97.05                    &             79.16         &    81.88                    &  89.37                       &              78.10          &   85.11                  &     22.19 &  3.43                   \\
MFN + AdaArcDistill     &                                   & 99.22                & 89.20                   & 92.47                    & 93.85                   & 85.62                   &\textbf{92.07 }                &   \textbf{71.33 }     &    \textbf{24.83}          \\ \hline
MFN (Student)           & \multirow{2}{*}{Casia-WebFace \cite{CASIAWbFace}}           &   98.97                   &         93.21              &    91.40                      &  91.27                      & 86.42                        &       92.25             &    77.46        &      57.75      \\
MFN + AdaArcDistill     & &  99.38                                 &   94.53             & 95.23                 &     94.57                &   90.35                &   \textbf{94.81}               &      \textbf{91.49}  & \textbf{86.25}                                 \\ \hline
\end{tabular}}
\caption{Evaluation results of MFN trained with two different datasets, IDiff-Face and CasiaWebFace. The teacher is trained on MS1MV2.}
\label{tab:dataset_study}
\vspace{-10mm}
\end{table*}

\textbf{Identity-disjoint teacher and student training datasets:}
Given that the class centers in our AdaDistill are calculated based on the exponential moving average of $f^t$ (Eq.\ref{eq:w}), the student and the teacher, unlike the solution presented in \cite{MarginKD,KD,ShrinkTeaNet}, do not necessarily need to share the same training dataset. 
To validate that, we trained standalone MFN models on Casia-WebFace \cite{CASIAWbFace} and IDiff-Face \cite{Boutros_2023_ICCV}, respectively using ArcFace loss. Then MFN models are trained again as a student using our AdaArcFace (Equations \ref{eq:arcdistill} \ref{eq:w} and \ref{eq:w_alpha} and $m1=0.45$), and the ResNet50 as the teacher (trained on MS1MV2).
Table \ref{tab:dataset_study} presents the verification accuracies of the standalone MFN and MFN with AdaArcDistill models.
As expected, our AdaArcDistill boosted the MFN verification accuracies under the considered evaluation benchmarks, even though the distillation used different data than the teacher's training data. 

\vspace{-5mm}
\subsection{Comparison with SOTA}
\vspace{-2mm}
We compare our AdaDistill with SOTA KD methods on 10 mainstream benchmarks described in Section \ref{sec:dataset}. 
The results of our AdaDistill are reported with two loss functions, noted as AdaArcDisill and AdaCosDistill. Our models are trained only on MS1MV2 \cite{guo2016ms,deng2019arcface} for fair comparisons with previous works.
Note that not all reported previous KD methods are designed for face recognition \cite{KD,DarkRank,SP,RKD,SFTN} or evaluated on the recent face benchmarks in their respective works \cite{FitNet,CCKD,TripletDistillation,SHKD}, and thus, we reported the results from \cite{ReFO,EKD,MarginKD,ShrinkTeaNet} when it was available.
The evaluation results of our AdaDistill and SOTA KD methods are reported in Table \ref{tab:sota} 
using evaluation metrics described in Section \ref{sec:metric}.
Several observations can be made from the reported results in Table \ref{tab:sota}. 
\begin{itemize}
\item  On the small verification benchmarks, LFW, CFP-FP, AgeDB-30, CA-LFW, and CP-LFW, our AdaArcDstill achived compatitive results to all SOTA techniques, exceeding them in many cases, with the best average accuracy of 95.43\% achieved by our AdaArcDstill.
\item On the large-scale IJB-C and IJB-B datasets, our AdaArcDstill and AdaCosDstill outperformed all SOTA methods. The best verification accuracies on IJB-C and IJB-B are 93.27\% and 91.21\% (TAR@FAR1e-4), respectively.
\item  Our AdaArcDstill and AdaCosDistill demonstrate superiority over the SOTA methods on the recent challenging ICCV21-MFR benchmarks, including all three testing sets, MR-all, Children, and Mask. 
\item  Our AdaArcDstill and AdaCosDistill are ranked among the top-performing methods on the MegaFace and MegaFace (R) benchmarks.
\item  Compared to the KD method aimed at transferring knowledge from angular space, MarginKD \cite{MarginKD} and ShrinkTeaNet \cite{ShrinkTeaNet}, our AdaArcDistill and AdaCosDistill outperformed these methods  \cite{MarginKD,ShrinkTeaNet} on all benchmarks.
\item  In comparison to the recent SOTA KD method (ReFO and ReFO+ CVPR2023 \cite{ReFO}) designed for FR, our AdaDistill clearly outperformed ReFO \cite{ReFO} on all large-scale challenging benchmarks, including IJB-C and ICCV21-MFR. As we discussed earlier, our AdaDistill achieved competitive results on MegFace, in comparison to ReFO \cite{ReFO} and other SOTA methods. On small evaluation benchmarks, our AdaDistill and ReFO achieved very competitive results. However, unlike ReFO \cite{ReFO}, our AdaDistill does not require student proxy or reverse distillation through multi-phases of student-teacher training.
    
\end{itemize}


\begin{table*}[t!]
\centering
\resizebox{\linewidth}{!}{%
\begin{tabular}{l|llllll|ll|ll|lll|llll}
\hline
\multirow{2}{*}{Method}                                 & \multirow{2}{*}{LFW} & \multirow{2}{*}{CFP-FP} & \multirow{2}{*}{AgeDB-30} & \multirow{2}{*}{CA-LFW} & \multirow{2}{*}{CP-LFW} & \multirow{2}{*}{Avg.} & \multicolumn{2}{l|}{IJB-C}        & \multicolumn{2}{l|}{IJB-B}       & \multicolumn{3}{l|}{ICCV21-MFR} & \multicolumn{4}{l}{MegaFace} \\ \cline{8-18} 
                                                        &                      &                         &                        &                        &                        &                       & 1e-4             & 1e-5           & 1e-4            & 1e-5           & MF-all    & Children   & Mask   & Id(R)  & Ver(R)  & Id  & Ver \\ \hline
IR50 (teacher)                                          & 99.80                & 97.63                   & 97.92                  & 96.05                  & 92.50                  & 96.78                 & 95.16            & 92.66          & 93.45           & 88.65           & 75.48     & 49.41         & 54.50 & 98.14 & \multicolumn{1}{l|}{98.34}  & 80.62          & 96.83    \\
MFN (student)                                           & 99.52                & 91.66                   & 95.82                  & 95.12                  & 87.93                  & 94.01                 & 89.13            & 81.65          & 87.07           & 74.63            & 53.43     & 24.71         & 27.90 & 90.91 & \multicolumn{1}{l|}{92.71}  & 75.52          & 90.80   \\ \hline
FitNet (arxiv14) \cite{FitNet}                          & 99.47                & 91.30                   & 96.18                  & 95.12                  & 88.30                  & 94.07                 & 87.76            & 73.71          & 86.35           & 70.19           & 54.46& 26.62  & 28.47 & 91.16 & \multicolumn{1}{l|}{92.34}  & 75.88          & 90.64     \\
KD (NIPSW14) \cite{KD}                                  & 99.50                & 91.71                   & 95.93                  & 95.03                  & 87.85                  & 94.00                 & 88.37            & 80.39          & 86.08           & 74.30            & 50.77& 26.36  & 25.74 & 90.40 & \multicolumn{1}{l|}{92.00}  & 75.81          & 90.07    \\
DarkRank (AAAI18) \cite{DarkRank}                       & 99.55                & 91.84                   & 95.60                  & 95.07                  & 87.77                  & 93.97                 & 89.28            & 81.62          & 86.76           & 73.75          & 56.82& 28.84  & 30.07 & 90.76 & \multicolumn{1}{l|}{92.41}  & 75.80          & 90.66    \\
SP (ICCV19) \cite{SP}                                   & 99.53                & 92.33                   & 96.17                  & 95.07                  & 88.45                  & 94.31                 & 88.43            & 78.13          & 86.34           & 72.85          & 54.44& 26.63  &  29.75 & 91.25 & \multicolumn{1}{l|}{92.41}  & 75.37          & 90.62   \\
CCKD (ICCV19) \cite{CCKD}                               & 99.47                & 91.90                   & 95.83                  & 95.22                  & 88.48                  & 94.18                 & 87.99            & 78.75          & 85.63           & 72.38           & 55.64& 27.65  & 30.22 & 91.17 & \multicolumn{1}{l|}{92.76}  & 75.73          & 90.63\\
RKD (CVPR19) \cite{RKD}                                 & 99.58                & 92.13                   & 96.18                  & 95.25                  & 87.97                  & 94.22                 & 89.65            & 83.21          & 87.27           & 75.17           & 53.92& 27.91  & 27.94 & 91.44 & \multicolumn{1}{l|}{92.92}  & 75.73          & 91.21   \\
ShrinkTeaNet (arxiv19) \cite{ShrinkTeaNet}              & 99.47                & 91.97                   & 96.00                  & 94.98                  & 88.52                  & 94.19                 & 87.80            & 79.78          & 85.31           & 75.23          & 55.28    & 27.73 & 30.24 & 90.73          & \multicolumn{1}{l|}{92.32}          & 75.55 & 90.56     \\
TripletDistillation (ICIP20) \cite{TripletDistillation} & 99.55                & 93.14                   & 95.53                  & 94.97                  & 88.03                  & 94.24                 & 84.57            & 76.65          & 81.88           & 70.51          &- &- &- & 86.52          & \multicolumn{1}{l|}{88.75}          & 71.93 & 91.35   \\
MarginKD (arxiv2020) \cite{MarginKD}                    & \textit{99.61}       & 92.01                   & 96.55                  & 95.13                  & 88.03                  & 94.27                 & 85.71            & 75.00          & 82.97           & 66.25        & 50.73& 25.14  & 28.54 & 91.70 & \multicolumn{1}{l|}{92.96}  & 76.34          & 91.31   \\
SFTN (NIPS21) \cite{SFTN}                               & 99.48                & 92.77                   & 96.30                  & -                      & -                      & -                     & 90.96            & 82.67          & -               & -              & 55.50& 28.51  & 29.66 & 91.69 & \multicolumn{1}{l|}{93.38}  & -              & -      \\
EKD (CVPR22) \cite{EKD}                                 & 99.60                & 94.33                   & 96.48                  & \textit{95.37}         & \textit{89.35}         & 95.03                 & 90.48            & 84.00          & 88.35           & 76.60          & 56.60& 28.95  & 32.14 & 91.02 & \multicolumn{1}{l|}{93.08}  & 75.54          & 91.42   \\
SH-KD (arxiv22) \cite{SHKD}                             & 99.47                & 94.67                   & 96.53                  & -                      & -                      & -                     & 91.75            & 85.76          & -               & -              & 57.69& 30.15  & 32.01 & \textbf{92.51} & \multicolumn{1}{l|}{\textbf{93.93}} & -     & -    \\
ReFO (CVPR23) \cite{ReFO}                               & 99.55                & 94.51                   & \textbf{96.92}         & -                      & -                      & -                     & 92.23            & 87.55          & -               & -             & 56.63& 33.36& 31.88 & 92.38 & \multicolumn{1}{l|}{93.80}  & -              & -     \\
ReFO+ (CVPR23) \cite{ReFO}                              & \textbf{99.65}       & 94.77                   & 96.42                  & -                      & -                      & -                     & 92.41            & 87.80          & -               & -               & \textit{59.17}& 32.80& 32.24& 92.41 & \multicolumn{1}{l|}{93.75}  & -              & -     \\ \hline
AdaArcDistill (ours)                                    & 99.58                & \textbf{95.24}          & \textit{96.78}         & \textbf{95.55}         & \textbf{90.02}         & \textbf{95.43}        & \textbf{93.27  } & \textbf{89.32} & \textbf{91.21 } & \textit{84.13} &  58.37    &  \textit{37.21}    & \textit{35.36} & 92.12 & \multicolumn{1}{l|}{93.32}  & \textbf{76.39} & \textbf{91.54}    \\
AdaCosDistill (ours)                                    & 99.50                & \textit{95.07}          & 96.60                  & 95.33                  & \textbf{90.02}         & \textit{95.30 }       & \textit{93.05}   & \textit{89.21} & \textit{90.93}  & \textbf{84.32}  &  \textbf{60.56 }   &  \textbf{39.94}    & \textbf{37.06} & 91.05 & \multicolumn{1}{l|}{92.68}  & 75.43& 90.83   \\ \hline
\end{tabular}
}
\caption{Evaluation results of our AdaDsitll (AdaArcDistill and AdaCosDistill) and various SOTA methods. The results are reported in (\%) on the five small benchmarks along with the average accuracy on these benchmarks. 
Our AdaDsitlls are among the top-performing methods on the small benchmarks with the best overall average accuracy. 
We also reported the results on the large-scale IJB-C and IJB-B benchmarks where our AdaDsitlls outperformed all SOTA methods. 
On the recent challenging ICCV21-MFR, our AdaDistill outperformed all SOTA methods and achieved very competitive results on MegaFace.
The best result on each dataset is in \textbf{bold} and the second is in \textit{italic}.
Results are reported using the evaluation metrics described in Section \ref{sec:metric}.
}
\label{tab:sota}
\vspace{-6mm}
\end{table*}

\vspace{-7mm}
\section{Limitation and social impact}
\vspace{-3mm}
Although our AdaDistill boosted the SOTA performances of compact FR, the verification accuracies are still comparably lower than the ones achieved by the computationally expensive SOTA FRs \cite{deng2019arcface,DBLP:conf/cvpr/HuangWT0SLLH20}, which might limit the deployability of such compact models in some scenarios, especially in critical use-cases that require the highest possible verification accuracies. The training of FR might involve collecting, sharing, or distributing authentic data that could be collected without proper user consent. 
Aiming at mitigating this issue, several recent works \cite{DBLP:journals/ivc/BoutrosSFD23,DBLP:conf/cvpr/Kim00023,Boutros_2023_ICCV} proposed the use of synthetic data to train standalone FR models. 
This work takes a step in this direction and proposes to train and evaluate a student within the KD framework using synthetically generated data.  

\vspace{-5mm}
\section{Conclusion}
\vspace{-3mm}
We proposed in this paper a novel Adaptive feature-based KD approach, AdaDistill, for deep face recognition. The proposed AdaDistill targets enhancing the verification accuracies of compact students while taking into consideration the relatively low capacity of shallow student architecture and the importance of the hard samples in the distillation process. 
This is achieved by building our distillation concept with the margin-penalty softmax while considering the class centers of the teacher.
This is achieved by adaptively distilling the class centers of the teacher to the student.
The adaptive nature is controlled by the progressing ability of the student to learn complex information over training iterations.
Through extensive experiments on mainstream face benchmarks, we demonstrated the effectiveness of our AdaDsitll and its superiority over the SOTA competitors, along with detailed experiments empirically proving our design choices.


\clearpage  

%
%
\bibliographystyle{splncs04}
\bibliography{main}
\end{document}